\documentclass[11pt]{elsarticle}
\usepackage{times}
\usepackage{booktabs}
\usepackage{multirow}
\usepackage{lineno,hyperref}
\usepackage{epsfig}
\usepackage{graphicx}
\usepackage{amsmath}
\usepackage{amssymb}
\usepackage{algorithm}
\usepackage{algorithmicx}
\usepackage{algpseudocode}
\usepackage{balance}
\usepackage{CJK}
\usepackage{acronym}
\usepackage{color}
\usepackage{colortbl}
\usepackage{array}
\usepackage{graphicx,bm}
\graphicspath{{Img/}}
\journal{Journal of Pattern Recognition}
\biboptions{numbers,sort&compress}
\modulolinenumbers[5]

\begin{document}
\begin{frontmatter}

\title{Deep Ranking Model by Large Adaptive Margin Learning for Person Re-identification}
\author[1]{Jiayun Wang}
\author[1]{Sanping Zhou}
\author[1]{Jinjun Wang\corref{cor1}}
\cortext[cor1]{Corresponding author:
  Tel.: +86-29-83395146;
  Fax: +86-29-83395175;}
\ead{jinjun@mail.xjtu.edu.cn.}

\author[1]{Qiqi Hou}
\address[1]{The institute of artificial intelligence and  robotic, Xi'an Jiaotong University, Xianning West Road No.28, Shaanxi, 710049, P.R. China}
\begin{abstract}
  Person re-identification aims to match images of the same person across disjoint camera views, which is a challenging problem in video surveillance. The major challenge of this task lies in how to preserve the similarity of the same person against large variations caused by complex backgrounds, mutual occlusions and different illuminations, while discriminating the different individuals. In this paper, we present a novel deep ranking model with feature learning and fusion by learning a large adaptive  margin between the intra-class distance and inter-class distance to solve the person re-identification problem. Specifically, we organize the training images into a batch of pairwise samples. 
  Treating these pairwise samples as inputs, we build a novel part-based deep convolutional neural network~(CNN) to learn the layered feature representations by preserving a large adaptive margin. As a result, the final learned model can effectively find out the matched target to the anchor image among a number of candidates in the gallery image set by learning discriminative and stable feature representations. Overcoming the weaknesses of conventional fixed-margin loss functions, our adaptive margin loss function is more appropriate for the dynamic feature space. On four benchmark datasets, PRID2011, Market1501, CUHK01 and 3DPeS, we extensively conduct comparative evaluations to demonstrate the advantages of the proposed method over the state-of-the-art approaches in person re-identification.
\end{abstract}

\end{frontmatter}

\section {Introduction}

\begin{figure}[!htb]
\footnotesize
\centering
\includegraphics[height = 5.0cm, width = 10.5cm]{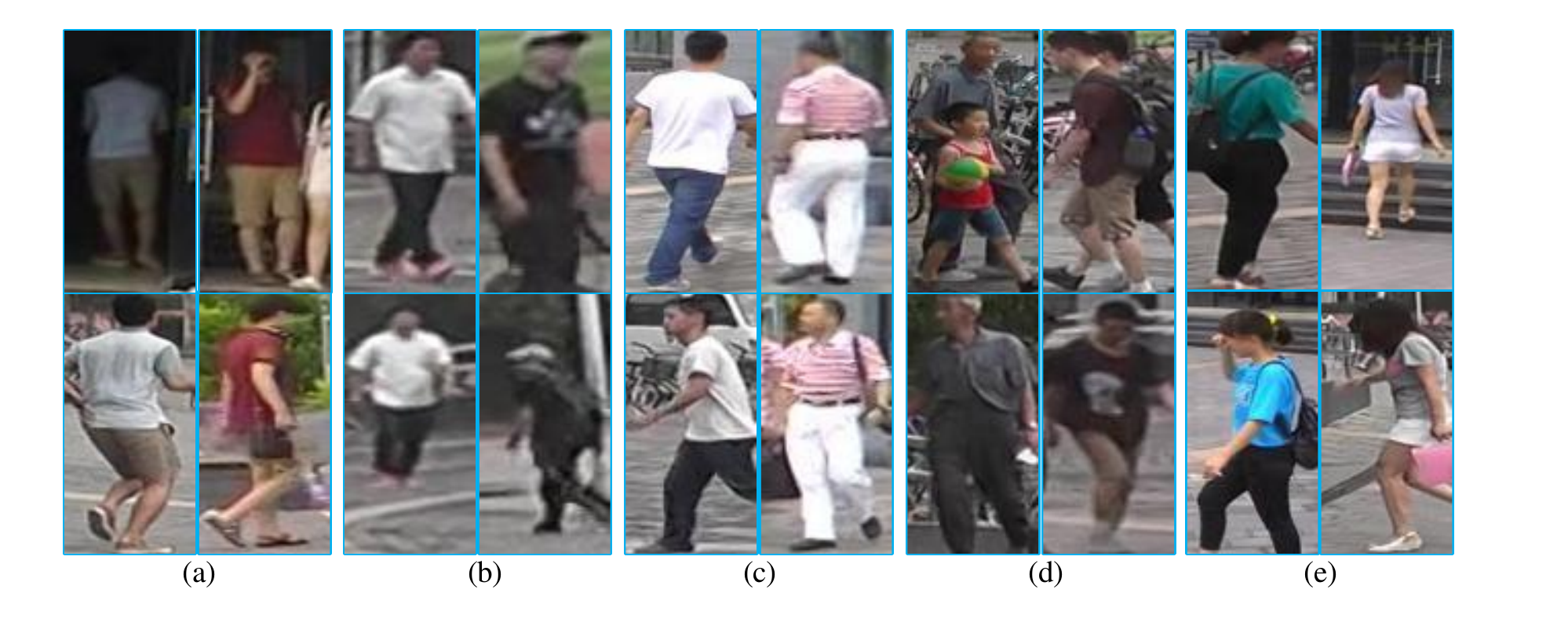}\\
\caption{The challenges to person re-identification in public space, in which the first row shows the training samples in camera view A and the
  second row represents the training samples in camera view B. Specially, (a) shows variations in lighting condition, (b) plots low resolution images, (c) illustrates variations of body pose, (d) denotes occlusions among pedestrians and (e) represents variations in view angle.}
\label{fig_1}
\end{figure}
Person re-identification~(re-id) is a fundamental task in video surveillance and has attracted much attention in the visual recognition community~\cite{Zhang_Wang_Wang:2015,Solera_Calderara_Cucchiara:2016,Hu_Tan_Wang:2004, varior2016gated}. Given one single shot or multiple shots of a target, it aims to match the same person among a set of gallery candidates captured from a disjoint camera view~\cite{Zhou_Wang_Hou:2016}. Since the matching results are ranked by the possibility of being the same identity, it is called "ranking" model. The problem is still extremely challenging due to large appearance variations caused by light conditions, view angles, mutual occlusions and body poses, as shown in Fig.~\ref{fig_1}. In addition, pedestrian images captured by the surveillance system are usually in low resolution, which makes many visual details, such as the facial components, indistinguishable. Therefore, different individuals may look similar in appearance.

To address these challenges, extensive work has been reported in the past few years, and can be roughly divided into two categories~\cite{Zhang_Xiang_Gong:2016,McLaughlin_Martinez_Miller:2016}: {\em feature representation} and {\em distance metric}. The feature representation based methods aim at constructing a discriminative visual descriptor that is robust to distinguish the different persons in disjoint camera views. Representative visual descriptors include the ensemble of local feature~(ELF)~\cite{Gray_Tao:2008}, local maximal occurrence~(LOMO)~\cite{Zhao_Ouyang_Wang:2014}, and local binary pattern~(LBP)~\cite{Xiong_Gou_Camps:2014}. The distance metric based methods aim at seeking a proper similarity measurement
by metric learning based on a group of labeled training data. Typical similarity measurements include the adaptive decision function~(LADF)~\cite{Li_Chang_Liang:2013}, large margin nearest neighbor~(LMNN)~\cite{Weinberger_Blitzer_Saul:2005} and pairwise constraint component analysis~(PCCA)~\cite{Davis_Kulis_Jain:2007}.

Recently, deep learning based methods are becoming increasingly popular in the person re-identification application~\cite{Ahmed_Jones_Marks:2015, Ding_Lin_Wang:2015, zhou2017point, wangcnnpack1,hubara2016binarized, xu2014large} because they can incorporate {\em feature representation} and {\em distance metric} into an integrated framework. The feature representation and distance metric are performed as two different components: (1) a deep CNN to extract feature representations from the input images, and (2) a designed metric to back-propagate gradients of the loss function. Benefiting from the powerful representation capability of the deep CNN, the deep learning based methods have achieved state-of-the-art performance on the benchmark datasets of person re-identification.

A number of deep learning based methods adopt the conventional contrastive loss or the triplet loss function to learn a fixed margin between the intra-class samples and inter-class samples, which do not suit the dynamic feature space well. The two loss functions also have some other drawbacks, as detailed in Section \ref{sec:comp}. Besides, as a result of lacking labeled training data, a person re-identification dataset is usually not huge in size, which highly limits the generalization ability of models on the testing data.  To alleviate these problems, we consider two notable aspects in our method: (1) the deep constraint should be adaptive to the dynamic feature space, and (2) the deep architecture should be more shallow and expressive, in order to cope with small or medium sized dataset. In this paper, we propose a novel deep ranking model with feature learning and fusion by learning a large adaptive  margin between the intra-class samples and inter-class samples for person re-identification. First, in order to generate discriminative feature representations for different individuals, we build a part-based deep CNN to learn feature representations from multiple perspectives of a pedestrian, in which different body parts are discriminatively learned in the convolutional layers and then fused in the fully connected layers. Secondly, we propose a novel distance metric to supervise the training of the deep CNN, in which the intra-class distance is minimized and the inter-class is maximized by adopting two large adaptive margin strategies, respectively.

In the following, we overview the main components of our method and summarize the advantages:
\begin{itemize}
  \item By organizing the training images into a batch of pairwise samples, we propose a novel distance metric, in which the intra-class distance is minimized and the inter-class distance is maximized by preserving a large adaptive margin between them. Compared with the existing approaches, our method can boost the ranking performance by adopting the adaptive margin strategy into the proposed distance metric.
  \item A novel part-based deep architecture is built to extract the discriminative feature representation of different body parts, which consists of three sub-networks, namely the global sub-network, local sub-network and fusion sub-network. Different body parts are first discriminately learned in the global sub-network and local sub-network, and then fused in the fusion sub-network. As a result, the finally learned feature representation is robust to the variations in light conditions, view angles, mutual occlusions and body poses.
  \item We conduct extensive experiments to evaluate various aspects of our method on the benchmark datasets, namely the PRID2011, Market1501, CUHK01 and 3DPeS, and the results outperform the state-of-the-art on all four datasets.
\end{itemize}

The rest of the paper is organized as follows. In Section~\ref{sec_rel}, we briefly review the related works. Section~\ref{sec_met} introduces the proposed large adaptive margin distance metric and the part-based deep neural network, followed by the optimization in Section~\ref{sec_opt}. Section~\ref{sec:generation} demonstrates the organization of training samples. The experimental results and result analysis are presented in Section~\ref{sec_exp}. Section~\ref{sec_con} concludes the work.

\section{Related Work}
\label{sec_rel}
There exists extensive work to tackle the person re-identification problem, focusing on feature representation, distance metric and deep learning. Below we present  some of the related work in terms of these three aspects.

{\bf Feature representation.} The feature representation based methods mainly focus on developing a discriminative visual descriptor, which is robust to the view angles, body poses, light conditions and mutual occlusions. For example, Zhao et al.~\cite{Zhao_Ouyang_Wang:2014} learned a mid-level filter from path cluster to achieve cross view invariance. Liao et al. \cite{Liao_Hu_Zhu:2015} constructed a feature descriptor, which analyzed this horizontal occurrence of local features and maximized the occurrence to make a stable representation against viewpoint changes. Ma et al.~\cite{Ma_Su_Jurie:2012} presented the person image via covariance descriptor which was robust to illumination change and background variations. Farenzena~\cite{Farenzena_Bazzani_Perina:2010} et al. augmented maximally stable color regions with histograms for person representation. Zhao et al.~\cite{Zhao_Ouyang_Wang:2013} learned the distinct salience feature to distinguish the matched person from others. Chen et al.~\cite{Cheng_Cristani_Stoppa:2011} employed a pre-learned pictorial structure model to localize the body parts more accurately. Wu et al.~\cite{Wu_Li_Radke:2015} introduced a viewpoint-invariant descriptor, which took the viewpoint of the human into account by using what they called a pose prior, learned from training data. Kviatkovsky et al.\cite{Koestinger_Hirzer_Wohlhart:2012} investigated the intra-distribution structure of color descriptor, which was invariant under certain illumination changes. Li et al.~\cite{Li_Wang:2013} matched person images observed in different camera views with complex cross-view transforms and applied it to person re-identification.

{\bf Distance metric.} The distance metric based methods aim to seek a proper similarity measurement, in which feature representations from the same person are closer than those from different ones. For example, Zheng et al.~\cite{Zheng_Gong_Xiang:2013} proposed a relative distance learning method from a probabilistic prospective. Mignon et al.\cite{Mignon_Jurie:2012} learned a distance metric from sparse pairwise similarity constraints. Pedagadi et al.~\cite{Pedagadi_Orwell_Velastin:2013} utilized local Fisher Discriminant Analysis~(LFDA) to map high dimensional features into a more discriminative low dimensional space. Xiong et al.~\cite{Xiong_Gou_Camps:2014} further extended the LFDA and several other metric learning methods by using kernel tricks and different regularizers. Nguyen et al.~\cite{Nguyen_Bai:2011} measured the similarity of face pairs through cosine similarity, which was closely related to the inner product similarity. Loy et al.~\cite{Loy_Liu_Gong:2013} modeled the person re-identification problem as an image retrieval task by considering the listwise similarity. Chen et al.~\cite{Chen_Yuan_Hua:2015} proposed a kernel based metric learning method to explore the nonlinearity relationship of samples in the feature space. Hirzer et al.~\cite{Hirzer_Roth:2012} learned a discriminative metric by using relaxed pairwise constraints.

{\bf Deep learning.} The deep learning based methods aim to incorporate feature extraction and metric learning into an integrated framework, in which feature representation can be learned under the supervision of one similarity metric.~\cite{yu2017deep, yu2017iprivacy, yu2015learning, sqe, DBLP1, DBLP2} For example, Li et al.~\cite{Li_Zhao_Xiao:2014} proposed a novel filter pairing neural network~(FPNN) to model body part displacements by using the patch matching layers to match the filter responses of local patches across views. Ahmed et al.~\cite{Ahmed_Jones_Marks:2015} proposed an improved deep learning framework which takes pairwise images as inputs, and outputs a similarity value indicating whether the two input images depict the same person or not. Xiao et al.~\cite{Xiao_Li_Ouyang:2016} proposed a domain guided dropout algorithm to improve the performance of deep CNN to extract robust feature representation for person re-identification. Yi et al.~\cite{Yi_Lei_Liao:2014} constructed a siamese neural network to learn pairwise similarity, and used body parts to train the model. Ding et al.~\cite{Ding_Lin_Wang:2015} applied the triplet loss to train the triplet deep framework for person re-identification. Wang et al.\cite{Wang_Zuo_Lin:2016} proposed a unified triplet and siamese deep architecture which can jointly extract single-image and cross-image feature representations. Zhao et al.~\cite{Zhao_Ouyang_Wang:2014} used the local patch matching method that learns the mid-level filters to get the local discriminative features for person re-identification.

\begin{figure*}[t!]
\centering
    \begin{tabular}{c}
        \includegraphics[height = 4.5cm, width = 12.2cm]{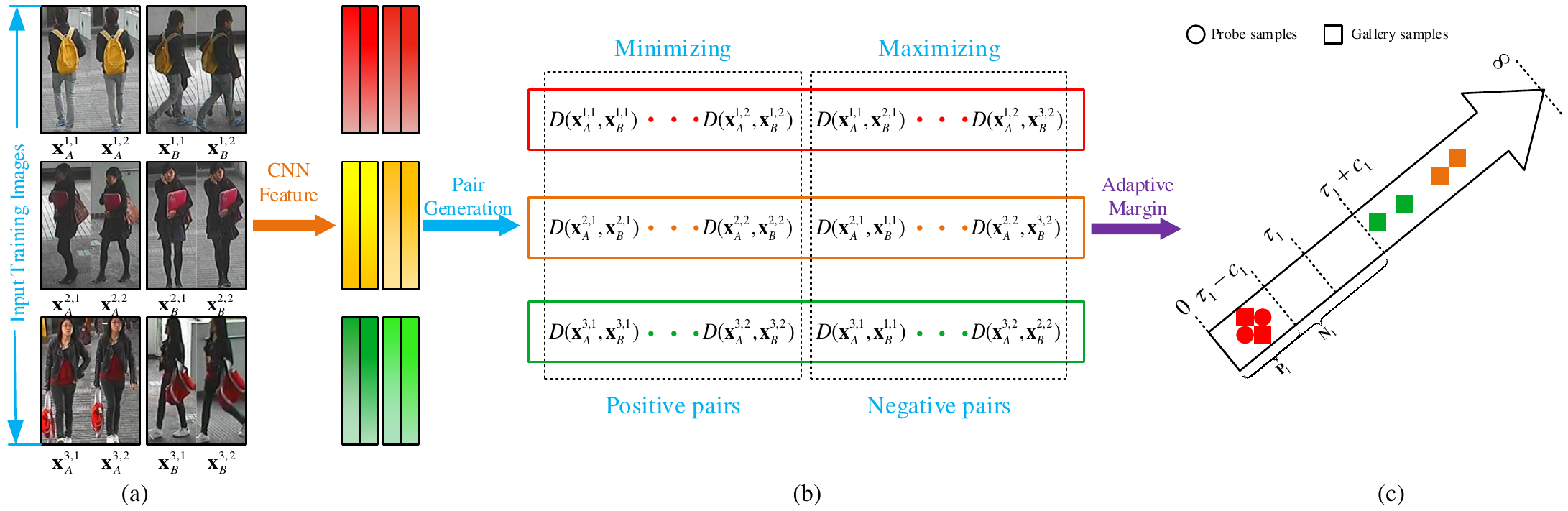}
    \end{tabular}
    \caption{The pipeline of the proposed methods. (a) shows the training samples feeding into the deep CNN; (b) the feature distance of positive pairs should be minimized, and that of negative pairs should be maximized; (c) shows the large adaptive margin strategy used in our method while minimizing the intra-class distance and maximizing the inter-class distance.}
    \label{fig_2}
\end{figure*}

\section{Our method}
\label{sec_met}
In this section, we first introduce the \textit{large adaptive margin loss function}, or adaptive loss, and then give the illustration of our \textit{deep neural network}. Fig.~\ref{fig_2} shows the basic pipeline of our method: the deep neural network is designed to extract feature representations of the input images, and the large adaptive margin metric is used to compute the loss and back-propagate the gradients. As a result, the optimized deep CNN can learn robust feature representations, in which the intra-class distance is smaller than the inter-class distance.

\subsection{Large adaptive margin}
\label{sec:lam}
Let $\mathbf{X} = \{\mathbf{X}_n\}_{n=1}^N$ be the set of training samples, where $\mathbf{X}_n = \{\mathbf{X}_A^n, \mathbf{X}_B^n\}$ denotes the $n^{th}$ person of training samples. $\mathbf{X}_n $ consists of data from both camera A and camera B. In each camera view, the training samples can be represented by $\mathbf{X}_A^n = \{\mathbf{x}_A^{n,i}\}_{i = 1}^{A_n}, \mathbf{X}_B^n = \{\mathbf{x}_B^{n,j}\}_{j = 1}^{B_n}$, where $\mathbf{x}_A^{n,i}, \mathbf{x}_B^{n,j}$ are the $i^{th}$ and $j^{th}$ raw input images of the $n^{th}$ person in both camera views, and $A_n, B_n$ denote the corresponding image number. Note that $\mathbf{X}$ denotes a set of images, while $\mathbf{x}$ only refers to one specific image. The goal of our deep model is to learn filter weights and biases that minimizes the ranking error from the output layer. A recursive function for an $M$-layer deep model can be formulated as follows:
\begin{equation}
\label{eq_1}
    \begin{aligned}
        \mathbf{X}_i^{(m)} &= \Psi_m(\mathbf{W}^{(m)}*\mathbf{X}_i^{(m-1)} + \mathbf{b}^{(m)}),\\
        i = 1, 2, \cdots&, N; m = 1, 2, \cdots, M; \mathbf{X}_i^{(0)} = \mathbf{X}_i.
    \end{aligned}
\end{equation}
where $\mathbf{W}^{(m)}$ denotes the filter weights of the $m^{th}$ layer, $\mathbf{b}^{(m)}$ refers to the corresponding biases, $*$ denotes the convolution operation, $\Psi_m(\cdot)$ is an element-wise non-linear activation function such as rectified linear unit~(ReLU), and $\mathbf{X}_i^{(m)}$ represents the feature maps generated at layer $m$ for sample $\mathbf{X}_i$. For similarity, we simplify the parameters of the neural network as a whole and define $\mathbf{W} = \{\mathbf{W}^{(1)}, \cdots, \mathbf{W}^{(M)}\}$ and $\mathbf{b} = \{\mathbf{b}^{(1)}, \cdots, \mathbf{b}^{(M)}\}$.

For each pair of raw input images $\mathbf{x}_A^{i,k}$ and $\mathbf{x}_B^{j,s}$, we represent them as $f(\mathbf{x}_A^{i,k})$ and $f(\mathbf{x}_B^{j,s})$ at the output layer for the subsequent similarity comparison. The pairwise distance metric can be measured by computing the squared Euclidean distance between the final feature representations, which is defined as follows:
\begin{equation}
\label{eq_2}
    \mathcal{D}(\mathbf{x}_A^{i,k}, \mathbf{x}_B^{j,s}) =  \|f(\mathbf{x}_A^{i,k}) -  f(\mathbf{x}_B^{j,s})\|_2^2
\end{equation}
The smaller $\mathcal{D}(\mathbf{x}_A^{i,k}, \mathbf{x}_B^{j,s})$ is, the more similar the two person images $\mathbf{x}_A^{i,k}$ and $\mathbf{x}_B^{j,s}$ are. Therefore, the definition formulates the person re-identification problem as a nearest neighbor search problem in the Euclidean space, which can be efficiently solved via similarity comparison algorithms.
\subsection{Adaptive margin loss function}
 The proposed adaptive margin loss function consists of two terms: the similarity comparison term and the regularization term, which are formulated as follows:
\begin{equation}
\label{eq_new}
 \mathrm{L} = \mathrm{L_S}(\mathbf{X}, \mathbf{W}, \mathbf{b})+\lambda R(\mathbf{W}, \mathbf{b})
\end{equation}
where $\mathrm{L_S}$ is the similarity comparison term, $\mathrm{R}$ represents the regularizer term, and $\lambda$ is the lagrangian multiplier. Given a set of pairwise training samples, the similarity comparison term supervises the training process by minimizing the intra-class distance or maximizing the inter-class distance, while preserving a large adaptive margin between them. The regularizer term is used to smooth the parameters of the deep model, so as to avoid overfitting. 

{\bf The similarity comparison term}. In order to exploit the stable and discriminative feature representations at the output layer of deep CNN, we expect that there is a large margin between the positive pairs and negative pairs. In a mini-batch, $\mathbf{x}_A^{i,k}$ denotes the $k^{th}$ capture of $i^{th}$ person in camera view A,  $\mathbf{X}_A^{g}$ denotes a set of  person captures in camera view A that has property $g$.
For the same person $\mathbf{x}_A^{i,k}$ and $\mathbf{x}_B^{i,s}$, the final distance $\mathcal{D}(\mathbf{x}_A^{i,k}, \mathbf{x}_B^{i,s})$ between $\mathbf{x}_A^{i,k}$ and $\mathbf{x}_B^{j,s}$ should be smaller than an adaptive up-margin $\mathcal{M}_p$. For different persons $\mathbf{x}_A^{i,k}$ and $\mathbf{x}_B^{j,s}$, the final distance $\mathcal{D}(\mathbf{x}_A^{i,k}, \mathbf{x}_B^{j,s})$ between $\mathbf{x}_A^{i,k}$ and $\mathbf{x}_B^{j,s}$ should be larger than an adaptive down-margin $\mathcal{M}_n$.  The formulation can be represented as follows:
\begin{equation}
\label{eq_3}
\begin{aligned}
    &\mathcal{D}(\mathbf{x}_A^{i,k}, \mathbf{x}_B^{j,s}) < \mathcal{M}_p(\mathbf{x}_A^{i,k}, \mathbf{X}_B^{g}), \hspace{0.1cm} i = j, \hspace{0.1cm} i \ne g;\\
    &\mathcal{D}(\mathbf{x}_A^{i,k}, \mathbf{x}_B^{j,s}) > \mathcal{M}_n(\mathbf{x}_A^{i,k}, \mathbf{X}_B^{g}), \hspace{0.1cm} i \ne j, \hspace{0.1cm} i = g.
\end{aligned}
\end{equation}
Obviously,  $\mathcal{M}_p$ is used to penalize the inter-class distances, and $\mathcal{M}_n$ is used to constrain the intra-class distances in the training process. $\mathcal{M}_p(\mathbf{x}_A^{i,k}, \mathbf{X}_B^{g}) ( \hspace{0.1cm} i \ne g)$ indicates that the up-margin is calculated based on the anchor and its negative pairs. For this purpose, we formulate the adaptive margins as follows:
\begin{equation}
\label{eq_4}
\begin{aligned}
    &\mathcal{M}_p = \frac{1}{\mu}(1 - \exp(-\mu d));\\
    &\mathcal{M}_n = \frac{1}{\gamma}\log(1 + \exp(\gamma s)).
\end{aligned}
\end{equation}
where the mean positive distance $s$ and mean negative distance $d$ are defined as follows:
\begin{equation}
\label{eq_av}
\begin{aligned}
    &s = \frac{1}{N}\sum\nolimits_{s = 1}^{N} \mathcal{D}(\mathbf{x}_A^{i,k}, \mathbf{x}_B^{j,s}),\; \text{if}\;i = j.\\
    &d = \frac{1}{N}\sum\nolimits_{d = 1}^{N} \mathcal{D}(\mathbf{x}_A^{i,k}, \mathbf{x}_B^{j,s}), \; \text{if}\;i \neq j.
\end{aligned}
\end{equation}

\begin{figure*}[t]
\centering
    \begin{tabular}{cc}
        \hspace{-0.1cm}
        \includegraphics[height = 3.5cm, width = 12.2cm]{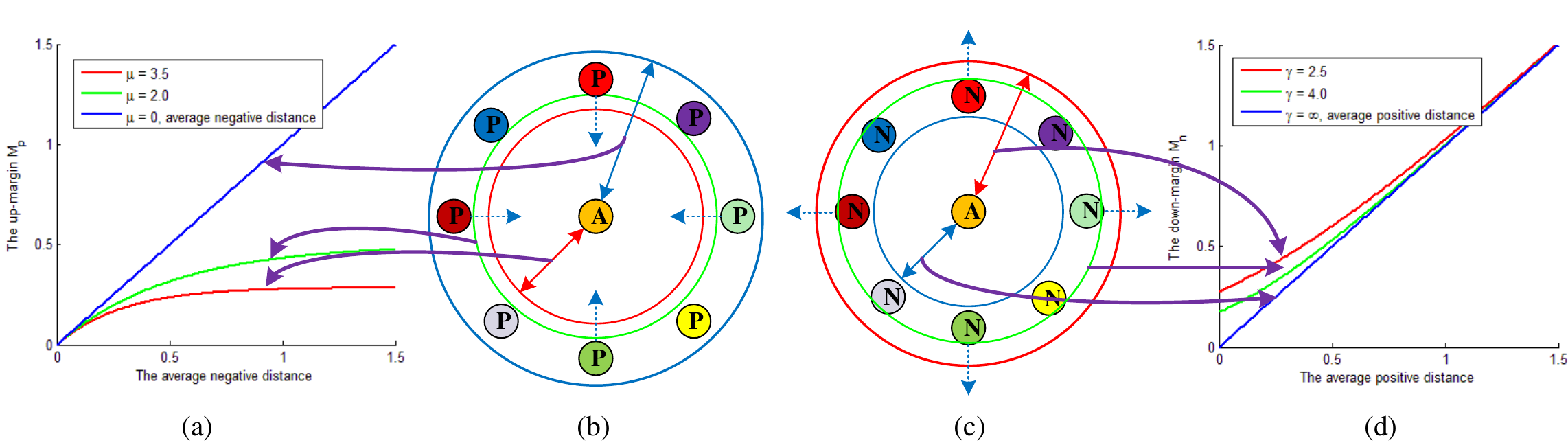}
    \end{tabular}
    \caption{The illustration of two different nonlinear mapping strategies for the adaptive margin, in which (a) and (b) show the margin constrained to $\mathcal{M}_p$, (c) and (d) represent the margin constrained to $\mathcal{M}_n$. Circles refers to the adaptive margins, the color of the circle is correspondent to different nonlinear mapping degree (controlled by hyper parameters $\mu, \gamma $) in (a) and (d). From the results, we can see that $\mathcal{M}_p$ will be penalized when the average negative distance is large and $\mathcal{M}_n$ will be compensated when the average positive distance is small. Besides, the degree of nonlinearization can be controlled by hyper parameters.}
    \label{fig_3}
\end{figure*}

We consider the adaptive margin $\mathcal{M}_p$ and $\mathcal{M}_n$ as a nonlinear mapping of the average distances $d$ and $s$. There are 2 advantages of the nonlinear mapping: 1) It is more suitable for the dynamic feature space. At the first several iterations of the training, the positive distance and the negative distance are similar. As depicted in Fig.~\ref{fig_3}, the nonlinear mapping penalizes the positive distance to get smaller, and the negative distance to get bigger, so that distinction between different identities can be achieved; when it comes to later iterations, as the difference between positive distance and negative distance gets more and more distinct, the nonlinear mapping effect diminishes significantly. Finally, $\mathcal{M}_p$ and $\mathcal{M}_n$ are close to the average positive distances and average negative distances, which can well avoid the over-fitting problem caused by the fixed margin strategy in the training process in earlier works. 2) The degree of the  nonlinear mapping can be tuned by the hyper parameters. As shown in Fig.~\ref{fig_3}, the larger $\mu$ gets, the more powerful the  nonlinear mapping becomes. Through tuning the 2 hyper parameters, our adaptive margin can adapt to different datasets. In short, the nonlinear mapping makes $\mathcal{M}_p$ get smaller than a moderate up-margin with the negative distances going up, and $\mathcal{M}_n$ get larger than a moderate down-margin with the positive distances going down.

Presetting $\mathcal{M}_p = \mathcal{M}_\tau - \mathcal{M}_c$ and $\mathcal{M}_n = \mathcal{M}_\tau + \mathcal{M}_c$, we can simplify the representation of constraint in Eq.\eqref{eq_3}, and the adaptive margin between the intra-class samples and inter-class sample can be enforced by using the following constraint:
\begin{equation}
\label{eq_5}
    \mathcal{M}_c - y_{ij}(\mathcal{M}_\tau - \mathcal{D}(\mathbf{x}_A^{i,k}, \mathbf{x}_B^{j,s})) < 0
\end{equation}
where $\mathcal{M}_\tau > \mathcal{M}_c$ and $y_{ij}$ represents the relationship between the $i^{th}$ and $j^{th}$ identities, which is defined as follows:
\begin{equation}
\label{eq_6}
y_{ij}= \left\{
\begin{aligned}
         1,& \hspace{0.2cm} \text{if} \hspace{0.2cm} i = j;\\
        -1,& \hspace{0.2cm} \text{else}.
\end{aligned} \right.
\end{equation}

By applying the adaptive margin constraint in Eq.~\eqref{eq_4} to each positive pair and negative pair in the training set, the hinge-like loss function of $\mathrm{L_S}$ can be formulated as follows:
\begin{equation}
\label{eq_7}
    \mathrm{L_S} = \sum\limits_{i,j = 1}^{N} \max\{\mathcal{M}_c - y_{ij}(\mathcal{M}_\tau - \mathcal{D}(\mathbf{x}_A^{i,k}, \mathbf{x}_B^{j,s})),0\}
\end{equation}

{\bf The regularization term} In order to smooth the parameters of the whole neural network, we define the following regularizer term, which can be formulated as follows:
\begin{equation}
\label{eq_8}
    \mathrm{R} = \sum\limits_{m = 1}^M \| \mathbf{W}^{(m)}\|_F^2 + \| \mathbf{b}^{(m)}\|_2^2
\end{equation}
where $\|\cdot\|_F^2$ denotes the Frobenius norm, and $\|\cdot\|_2^2$ represents the Euclid norm.

Putting Eq.~\eqref{eq_7} and Eq.~\eqref{eq_8} into Eq.~\eqref{eq_new}, the loss function of our method can be formulated as follows:
\begin{equation}
\label{eq_9}
     \mathrm{L} = \sum\limits_{i,j = 1}^{N} \max\{\mathcal{M}_c - y_{ij}(\mathcal{M}_\tau - \mathcal{D}(\mathbf{x}_A^{i,k}, \mathbf{x}_B^{j,s})),0\} + \lambda (\sum\limits_{m = 1}^M \| \mathbf{W}^{(m)}\|_F^2 + \| \mathbf{b}^{(m)}\|_2^2)
\end{equation}
Using the large adaptive margin algorithm to compute the loss and the back propagation algorithm to update the deep parameters, the final learned deep ranking model can well distinguish the different individuals by extracting the discriminative feature representations.

\subsection{Comparison with two fixed margin loss functions}
\label{sec:comp}
\begin{figure}[t!]
\centering
\includegraphics[width=1.0\textwidth]{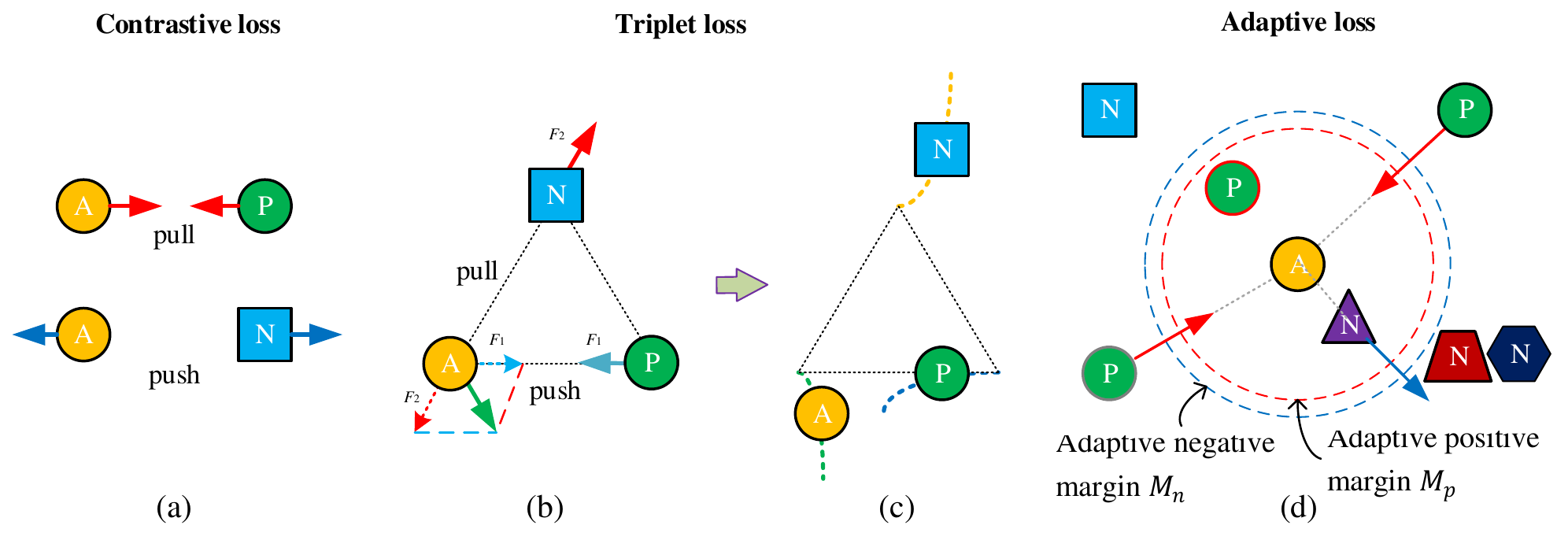}
\caption{Comparison of our method with two fixed margin approaches, in which (a) shows that the contrastive loss function penalizes the positive distances and negative distances with an optimal gradient back-propagation direction but a fixed margin; (b) and (c) show the triplet loss function minimizes the relative distances between the intra-class samples and inter-class samples with a fixed margin and suboptimal gradient back-propagation direction; (d) shows our adaptive margin loss function penalizes the positive distances and negative distances with an adaptive margin and optimal gradient back-propagation direction.}
 \label{fig:prob}
\end{figure}

In order to illustrate how the adaptive margin strategy works, we compare our algorithm with two fixed margin loss functions, namely \emph{the contrastive loss function} and \emph{the triplet loss function}. From Fig. \ref{fig:prob}, we can conclude that: 1) \emph{The contrastive loss function} has the optimal updating direction of back-propagation, while it penalizes the intra-class distance to zeros and the inter-class distances bigger than a fixed positive margin. 2) \textit{The triplet loss function} minimizes the relative distance between the intra-class samples and the inter-class samples, which also applies the fixed margin strategy. A further critical restriction is that the updating direction of back-propagation deduced by the loss function is not perfect, which can easily make the intra-class distance going up in further iterations. Our \emph{adaptive margin loss function} aims to penalize an adaptive up-margin between the intra-class samples and an adaptive down-margin between the inter-class samples, which can overcome the drawbacks of \emph{the contrastive loss function} in the margin strategy and \emph{the triplet loss function} in both the margin strategy and gradient back-propagation. In most of the deep learning based person re-identification methods, \emph{the triplet loss function} has been proved to be more effective than \emph{the contrastive loss function}, because large number of triplets can be generated in the training process\footnote{Suppose one data set contains $M$ persons from two disjoint camera views, and each person has $N$ images in each camera view, therefore the number of different triplets is $M(M-1)N^3$, while the number of different positive pairs and negative pairs is $MN^2 + M(M-1)N^2$.}. Benefiting from the adaptive margin strategy, our method can outperform \emph{the triplet loss function} in training the deep CNN. The underlying mechanism is that both the positive distances and negative distances are small at the early iterations, and the positive distances will increase slowly compared to the negative distances with further iterations. Therefore, we will gradually increase the upper-margin $M_p$ to avoid the over-fitting problem by excessively penalizing the positive distances. Meanwhile, we also gradually increase the down-margin $M_n$ to enhance the discrimination of positive pairs from the negative ones. This mechanism can be simply realized by using our adaptive margin strategy, in which we use the average negative distance to model the upper-margin for positive pairs and the average positive distance to model the down-margin for negative pairs. Therefore, the updating of $M_p$ and $M_n$ can jointly consider the variations of negative distance and positive distance, respectively.

\subsection{Deep neural network}
\label{sec:nn}
\begin{figure*}[t!]
\centering
    \begin{tabular}{cc}
        \hspace{-0.1cm}
        \includegraphics[height = 6.2cm, width = 12.0cm]{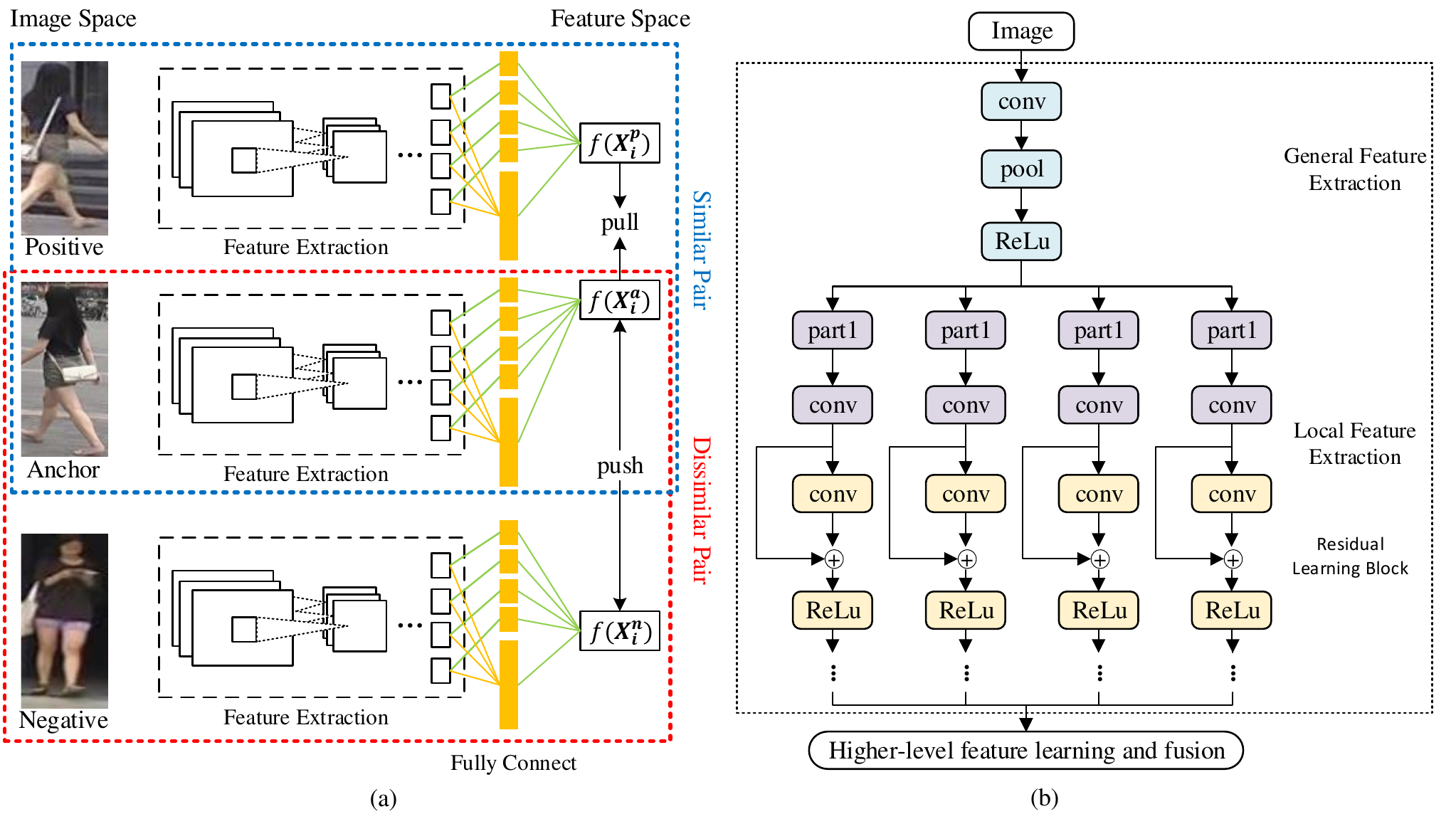}
    \end{tabular}
    \caption{The deep feature learning and fusion neural network. Firstly, we organize the input images into the positive pairs and negative pairs. Secondly, we apply the global feature learning, local feature learning and residual learning in the convolutional layers. Note that the number of residual learning blocks in local feature learning can vary in different datasets. Thirdly, we implement the local feature learning and fusion in the fully connected layers.  Finally, the concatenated feature representations are fed into the adaptive margin metric for similarity comparison.}
    \label{fig_4}
\end{figure*}

The proposed adaptive margin metric is combined with our part-based deep CNN to implement an end-to-end feature learning and fusion for person re-identification. The whole learning process is illustrated in Fig.~\ref{fig_4}, in which we firstly organize the training samples into the positive pairs and negative pairs, then we apply the adaptive metric to supervise the leaning of neural network in a Siamese framework. The part-based deep neural network is consisted of three parts, which is small in size but effective in performance. In the following paragraphs, we will introduce each part of the deep neural network in detail.

{\bf Global Feature Learning} The first part of our network is used for global feature learning, which consists a convolutional layer and max pooling layer. They are used to extract the low-level features of the input images, so as to provide multi-level feature representations to be discriminatively learned in the following local sub-network. The input images are in size of $230\times80\times 3$, and are firstly passed through $64$ learned filters of size $7\times7\times3$. Then, the resulting feature maps are passed through a max pooling kernel of size $3\times3\times3$ with stride $3$. Finally, these feature maps are passed through a rectified linear unit~(ReLU).

{\bf Local Feature Learning} The second part of our network is used for local feature learning, which consists of four teams of convolutional layers and max pooling layers. We firstly divide the input feature maps into four equal horizontal patches across the height channel, which introduces $4\times64$ local feature maps of different body parts. Then, we pass each local feature maps through two convolutional layers, and both of them have $32$ learned filters of size $3\times3$. Furthermore, the outputs of the first local convolutional layer are summarized with the outputs of the second local convolutional layer using element-wise operation, which we called residual block. Note that the number of residual learning block can be different for different dataset. The influence of the number of the residual blocks is discussed in Section \ref{sec:deeper}. Afterwards, the resulting feature maps are passed through max pooling kernels of size $3\times3$ with stride 1. Finally, we add a ReLU after each max pooling layer. In order to learn the feature representations of different body parts discriminately, we do not share the parameters among the four paths of convolutional layers.

{\bf Feature Learning and Fusion} The third part of our network is used for feature learning and fusion sub-network, which consists  of four teams of fully connected layers. Firstly, the local feature maps of different body parts are discriminately learned by following two fully connected layers in each path. The dimension of the fully connected layer is $100$ and a ReLU is added between them. Then, the discriminately learned local feature representations of the first four fully connected layers are concatenated to be summarized by adding another fully connected layer, whose dimension is 400. Finally, the resulting feature representation is further concatenated with the outputs of the second four fully connected layers to generate 800 dimensional final feature representations. Similarly, we do not share the parameters among the four fully connected layers to perserve the discrimination of feature representations of different body parts.

\section{Optimization}
\label{sec_opt}
We use the gradient back-propagation method to optimize the parameters of the deep CNN, which is carried out in the mini-batch manner. Therefore, we need to calculate the gradients of the loss function with respect to the feature representations at the output layers. For simplicity, we consider the parameters in the network as a whole and define $\mathbf{\Omega}^{(m)} = [\mathbf{W}^{(m)}, \mathbf{b}^{(m)}]$, and $\mathbf{\Omega} = \{\mathbf{\Omega}^{(1)}, \dots, \mathbf{\Omega}^{(M)}\}$.

In order to employ the back-propagation algorithm to optimize the network parameters, we compute the partial derivative of the loss function as follows:
\begin{equation}
\label{eq_10}
    \frac{\partial \mathrm{L}}{\partial \mathbf{\Omega}} = \sum\limits_{i,j = 1}^N \frac{\partial \mathrm{L}}{\partial f(\mathbf{x}_A^{i,k})} \cdot \frac{\partial f(\mathbf{x}_A^{i,k})}{\partial \mathbf{\Omega}} + \frac{\partial \mathrm{L}}{\partial f(\mathbf{x}_B^{j,s})} \cdot \frac{\partial f(\mathbf{x}_B^{j,s})}{\partial \mathbf{\Omega}} + 2 \lambda \sum\limits_{k = 1}^M \mathbf{\Omega}^{(k)}
\end{equation}
where the first two terms represent the gradient of the similarity comparison term, and the third term is the gradient of the regularization term.

For simplicity, we define $\mathcal{R} = \mathcal{M}_c - y_{ij}(\mathcal{M}_\tau - \mathcal{D}(\mathbf{x}_A^{i,k}, \mathbf{x}_B^{j,s}))$, then ${\partial \mathrm{L}}/{\partial f(\mathbf{x}_A^{i,k})}$ and ${\partial \mathrm{L}}/{\partial f(\mathbf{x}_B^{j,s})}$ can be formulated as follows:
\begin{equation}
\label{eq_11}
     \frac{\partial \mathrm{L}}{\partial f(\mathbf{x}_A^{i,k})} = \left\{
     \begin{aligned}
     -y_{i,j}(\mathcal{M}_{\tau} - 2(&f(\mathbf{x}_A^{i,k}) - f(\mathbf{x}_B^{j,s}))), \hspace{0.1cm} if \hspace{0.2cm} \mathcal{R} > 0;\\
     & 0, \hspace{1.8cm} else.
     \end{aligned}\right.
\end{equation}

\begin{equation}
\label{eq_12}
     \frac{\partial \mathrm{L}}{\partial f(\mathbf{x}_B^{j,s})} = \left\{
     \begin{aligned}
     -y_{i,j}(\mathcal{M}_{\tau} + 2(&f(\mathbf{x}_A^{i,k}) - f(\mathbf{x}_B^{j,s}))), \hspace{0.1cm} if \hspace{0.2cm} \mathcal{R} > 0;\\
     & 0, \hspace{1.8cm} else.
     \end{aligned}\right.
\end{equation}

The gradients of the adaptive margin metric can be easily calculated given the values of $f(\mathbf{x}_A^{i,k}), f(\mathbf{x}_B^{j,s})$ and ${\partial f(\mathbf{x}_A^{i,k})}/{\partial \mathbf{\Omega}}, {\partial f(\mathbf{x}_A^{i,k})}/{\partial \mathbf{\Omega}}$ in each mini-batch, in which they can be obtained by separately running the forward and backward propagation for all the positive pairs and negative pairs. Since the algorithm needs to go through all the pairwise units to accumulate the gradients in each iteration, we call it the adaptive margin gradient descent algorithm. We
show the overall process in Algorithm~\ref{alg}.
\begin{algorithm}
\caption{The adaptive margin gradient descent algorithm}
\label{alg}
\begin{algorithmic}[1]
\State{\textbf{Input:}} Training samples $\mathbf{X}$, learning rate $\omega$, maximum iterations $H$, adaptive margin parameters $\mu$ and $\gamma$, weight parameter $\lambda$, updating rate $\eta$.
\State{\textbf{Output:}} The deep parameters $\mathbf{\Omega}$.
\While {${h < H}$}
\State Calculate the output feature representations of $f(\mathbf{x}_A^{i,a})$ and $f(\mathbf{x}_B^{j,s})$ for all the positive pairs and negative pairs in a mini-batch by forward propagation;
\State Compute the adaptive margin $\mathcal{M}_p$ and $\mathcal{M}_n$ according to Eq.~\eqref{eq_4};
\State Compute the ${\partial f(\mathbf{x}_A^{i,k})}/{\partial \mathbf{\Omega}}, {\partial f(\mathbf{x}_A^{i,k})}/{\partial \mathbf{\Omega}}$ according to Eq.~\eqref{eq_11} and Eq.~\eqref{eq_12};
\State Compute the $\frac{\partial \mathrm{L}}{\partial \mathbf{\Omega }}$ according to Eq.~\eqref{eq_10};
\State Update the deep parameters $\mathbf{\Omega}_{h+1} = \mathbf{\Omega}_h - {\tau_h}\frac{\partial \mathrm{L}}{\partial \mathbf{\Omega}_h}$ and $h \leftarrow h + 1$.
\EndWhile
\end{algorithmic}
\end{algorithm}

\section{Mini-batch Generation}
\label{sec:generation}

\begin{figure*}[!htb]
\centering
    \begin{tabular}{cc}
        \hspace{-0.1cm}
        \includegraphics[height = 6.2cm, width = 12.0cm]{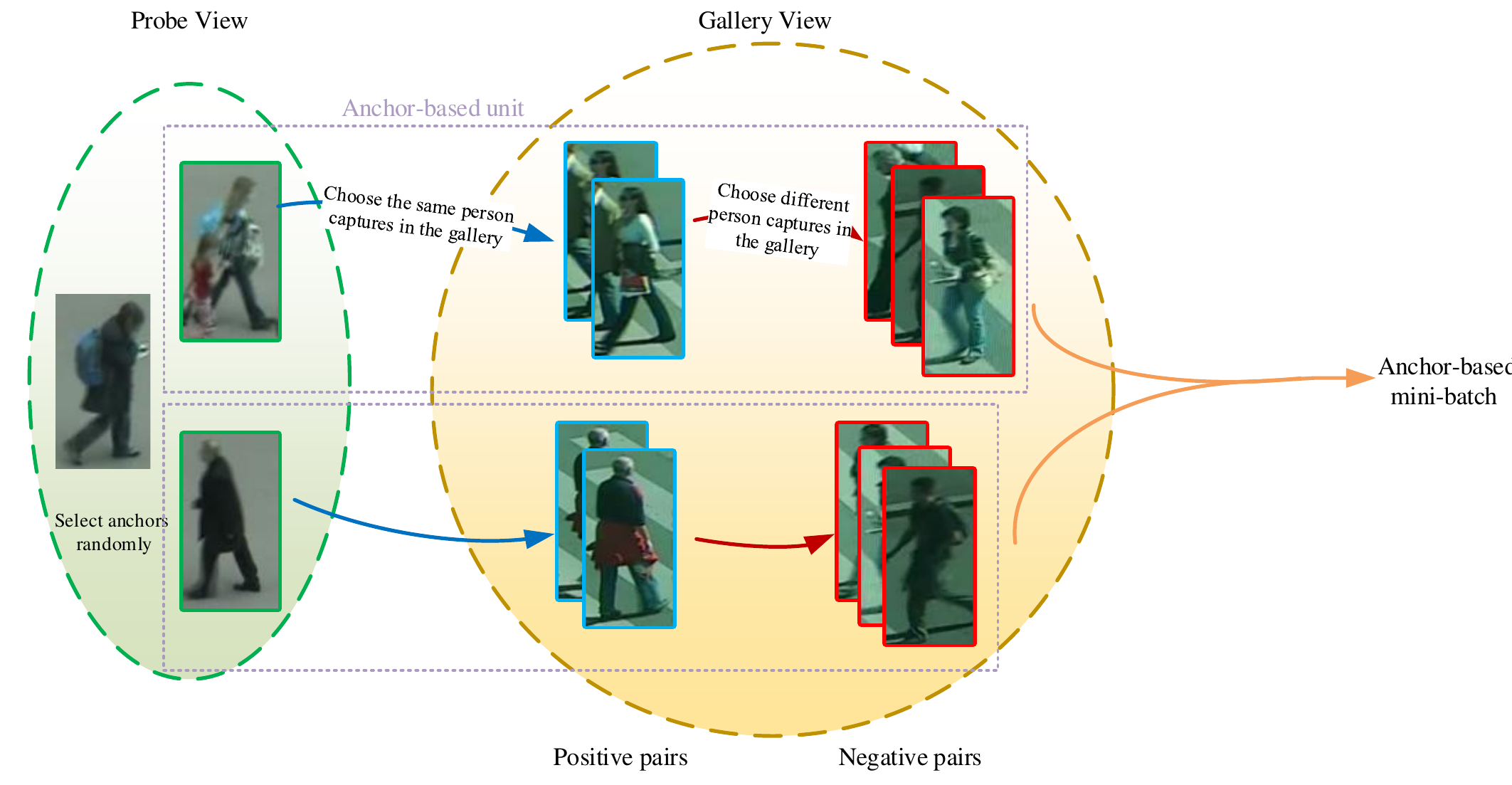}
    \end{tabular}
    \caption{Illustration of the anchor-based minibatch generation. Firstly, we randomly find a subset of the captures of persons in the probe view, and each image is called anchor. Secondly, we randomly find $M$ corresponding positives and $N$ negatives in the gallery, respectively. Finally, the generated positive pairs and negative pairs are fed into the neural network in a min-batch manner.}
    \label{fig:anchor-based}
\end{figure*}

\begin{algorithm}[h]
\caption{Learning features of the training set in the mini-batch mode} 
\hspace*{0.02in} {\bf Input:} 
\\\hspace*{12pt} Labeled training images$\{I_i\}$.\\
\hspace*{0.02in} {\bf Output:} 
\\\hspace*{12pt}The network parameters $\bm{\Omega}$.
\label{alg2}
\begin{algorithmic}[1]
    \While{$n<N$}
    \State Randomly select $A$ anchors from the training set;
    \For{all anchors}
    \State Randomly select $M$ positives and $N$ negatives and form the anchor-based unit.
    \EndFor
    \State Form the mini-batch and $n\leftarrow n+1$.
    \EndWhile
    \For{all mini-batches}
    \State Update $\bm{\Omega}$ according to \hyperref[alg]{Algorithm \ref{alg}};
    \EndFor
\end{algorithmic}
\end{algorithm}

As the person re-identification datasets usually contains hundreds of or even thousands of pedestrians, it is impossible to load all the positive pairs and negative pairs due to the memory limit and computational speed. Therefore, it is necessary to train the neural network in the mini-batch mode. To be more specific, we only use a small set of anchor-based units to train the neural network in each iteration. 

The basic pipeline of anchor-based mini-batch generation is shown in Fig. \ref{fig:anchor-based}. Given a labeled training set, we randomly find a subset of images in the probe view, where each image is called an \emph{anchor}. Then, we randomly find $M$ corresponding positives and $N$ negatives for each anchor in the gallery view. Finally, the generated positive pairs and negative pairs are fed into the neural network in a minbatch manner. By repeating the steps above several times, we can obtain a large number of possible anchor-based units without overburdening the hardware, because random selection mechanism guarantees huge possibilities of the input combination. Algorithm \ref{alg2} shows the complete batch training process.
\section{Experiments}
\label{sec_exp}
\subsection{Datasets and Settings}
We evaluate the proposed method on four widely used benchmark datasets, namely PRID2011~\cite{Hirzer_Beleznai_Roth:2011}, Market1501~\cite{Zheng_Shen_Tian:2015}, CUHK01~\cite{Li_Wang:2013} and 3DPeS~\cite{Baltieri_Vezzani_Cucchiara:2011}. Each of them has at
least 2 image for each person from each camera view.

{\bf PRID2011} is specially designed for video-based person re-identification problem, including 749 persons captured by two disjoint cameras, with sequences lengths of 5 to 675
frames. We evaluated the proposed methods on two settings: with temporal information, and without temporal information. The evaluation protocol follows~\cite{You_Wu_Li:2016} and \cite{rnncvpr}, respectively. 

{\bf Market1501} is a large scale person re-identification dataset, containing 32668 images of 1501 individuals. Each individual is captured by two to six cameras. We use the provided fixed training and test set, under the multi-query evaluation settings in~\cite{Zhang_Xiang_Gong:2016}.

{\bf CUHK01} contains 971 persons captured from two camera views in a campus environment, and there are two images for each person from every camera view. We utilize the same protocol with~\cite{Wang_Zuo_Lin:2016}, where 871 person images are used for training and the rest for testing.

{\bf 3DPeS} has 1011 images of 192 persons captured from 8 outdoor cameras with significantly different viewpoints. The number of images for each person varies from 2 to 26. We utilize the same protocol as in~\cite{Chen_Yuan_Chen:2016}, where half of the persons are used for training and the rest for testing.

All the results are evaluated under cumulative matching characteristic~(CMC)~\cite{Gray_Brennan_Tao:2007}, which is an estimation of finding the corrected match in the top $n$ match. Final performance is averaged over ten random repeats of the process. Specially for Market1501, we also report mAP result.

{\bf Parameter setting.} The weights are initialized from two zero-mean Gaussian distribution with the standard deviations from $0.01$ to $0.001$, respectively. The bias terms are set to $0$. The learning rate $\omega = 0.01$, the updating rate $\eta = 0.001$, the weight parameter $\alpha = 0.01$.


\subsection{Results}
\label{sec:expresults}


Comprehensive evaluation is conducted on the proposed method and compared with the state-of-the-art methods. For small-sized and medium-sized datasets, PRID2011, CUHK01 and 3DPeS, the network only adopts 1 residual learning block for each body part; for large scale datasets Market1501, we use 4 residual learning blocks for each body part, as well as transfer learning strategy. 


\textbf{PRID2011:} We evaluate the algorithm performance with and without temporal information. Table. \ref{table:PRID} show that the proposed method has the best performance in all top-1 to top-50 identification result. Without using temporal information, our method achieves 57.8\% top-1 accuracy, which outperforms the state-of-the-art method without temporal information discriminative null space method (DNS) \cite{Zhang_Xiang_Gong:2016} it by a large margin (57.8\% vs 40.9\%) in top-1 accuracy. 
When using temporal information and recurrent network \cite{rnncvpr}, our method improves the top-1 accuracy by 3.3 percent, compared to state-of-the-art method with temporal information~\cite{rnncvpr} .

\begin{table}[!hbp]
\centering
 \caption{ \label{table:PRID}
Identification rates (\%) of different methods on PRID2011 dataset}
\begin{tabular}{|c || c  c c c|}
\hline

Method & $r=1$ &  $r=10$  & $r=20$ & $r=50$ \\
\hline
DeepM \cite{Yi2014Deep}&17.9  &45.9 &55.4 &74.1\\
KISSME \cite{f21}&15.0  &39.0 &52.0 &68.0\\
Mahalanobis \cite{Hirzer_Beleznai_Roth:2011}&16.0 &41.0 &51.0 &64.0\\
EIML \cite{EIML}&16.0 &39.0 &51.0 &68.0\\
ITML \cite{Davis_Kulis_Jain:2007}&12.0&36.0&47.0&64.0\\
LMNN \cite{Weinberger_Blitzer_Saul:2005}&10.0 &30.0 &42.0 &59.0\\
DNS \cite{Zhang_Xiang_Gong:2016} &40.9  &73.2 &81.0 &90.4\\
Wu's~\cite{wu2016deep} \cite{Zhang_Xiang_Gong:2016} &69.0  &93.2 &96.4 & \\
RCN \cite{rnncvpr}  &70  &95 &97 & \\
\hline
Ours  &57.8  &90.3 &97.7 & 100 \\
Ours + RCN  &\textbf{73.3}  & \textbf{97.5} & \textbf{98.3} &\textbf{100} \\
\hline
\end{tabular}
\end{table}

\textbf{Market1501:}  
As a newly proposed large scale person re-identification dataset, the best performance was obtained by Geng et al. \cite{geng2016deep}. As shown in Table. \ref{table:Market}, the proposed method achieves the state-of-the-art top-1 identification rate at 89.5\%, and mAP at 74.1.

\begin{table}[!hbp]
\centering
 \caption{ \label{table:Market}
 Top-1 identification rate and mAP of various methods on Market 1501 dataset}
\begin{tabular}{|c || c c|}
\hline

Method & $r=1$ & mAP \\
\hline
BoW \cite{Zheng_Shen_Tian:2015}&  34.4& 14.1\\
Baseline(+Mahalanobis) \cite{Zheng_Shen_Tian:2015}  &36.8&  15.1\\
Baseline(+KISSME\cite{P1})  &42.7&  19.6\\
SCSP \cite{Chen_Yuan_Chen:2016}&  51.9  &26.4\\
XQDA(LOMO) \cite{Mignon_Jurie:2012} &54.1&  28.4\\
S-LSTM \cite{M15011}& 61.6  &35.3\\
DNS(LOMO) \cite{Zhang_Xiang_Gong:2016}  &68.0 &41.9\\

Gated S-CNN \cite{varior2016gated}  &76.0 &48.5\\

Zhong's \cite{zhong2017re}  &77.1 &63.6\\

Zheng's \cite{zheng2016discriminatively}  &85.8 &70.3\\
Geng's \cite{geng2016deep}  &\textbf{89.6} &73.8\\
\hline
Ours  &\textbf{89.5}  & \textbf{74.1} \\
\hline
\end{tabular}
\end{table}

\textbf{CUHK01:} 
From the results in Table \ref{table:c01}, the previous best top-1 accuracy was achieved by  Ejaz \emph{et al.} \cite{Ahmed_Jones_Marks:2015} at 65.0\%. Our method obtains 71.90\% top-1 accuracy, outperforming all other methods. Besides, top-1 to top-20 identification rates outperform the state-of-the-art-method by a large margin.

\begin{table}[!hbp]
\centering
 \caption{ \label{table:c01}
Identification rates (\%) of different methods on CUHK01 dataset}
\begin{tabular}{|c || c  c c c|}
\hline

Method & $r=1$ &  $r=5$  & $r=10$ & $r=20$ \\
\hline
Ejaz's \cite{Ahmed_Jones_Marks:2015}&65.0 &87.0 &94.0 &97.0\\
FPNN \cite{Li_Zhao_Xiao:2014}&20.7  &51.0 &67.0 &82.5\\
KISSME \cite{P1}&14.2 &36.0 &51.5 &69.0\\
LDM \cite{P4}&13.5  &36.0 &51.4 &70.5\\
RANK \cite{P5}&10.4&30.0&46.0&62.0\\
eSDC \cite{Zhao_Ouyang_Wang:2013}&8.8 &24.0 &36.0 &52.8\\
LMNN \cite{Weinberger_Blitzer_Saul:2005} &7.3 &19.8 &30.5 &48.0\\
ITML \cite{Davis_Kulis_Jain:2007} & 5.5 & 23.5& 35.5 & 50.8\\
\hline
Ours  &\textbf{71.9}  & \textbf{91.8} & \textbf{95.8} &\textbf{97.2} \\
\hline
\end{tabular}
\end{table}

\textbf{3DPeS:} 
The proposed method obtains 58.3\% top-1 accuracy, outperforming all the other methods, as presented in Table \ref{table:3dpes}. However, kLFDA \cite{Xiong_Gou_Camps:2014} achieves the highest top-5 identification rate and ME \cite{Chen_Yuan_Chen:2016} achieves the highest top-20 identification rate. Yet it is worth noting that the margin of these rates between the state-of-the-art method and our method is relatively small. 

\begin{table}[!hbp]
\centering
 \caption{ \label{table:3dpes}
Identification rates (\%) of different methods on 3DPeS dataset}
\begin{tabular}{|c || c  c c|}
\hline

Method & $r=1$ &   $r=5$ & $r=20$ \\
\hline
LF \cite{Pedagadi_Orwell_Velastin:2013}&45.5  &69.2   &89.06\\
ME \cite{3dpes3}&53.3 &76.8   &\textbf{92.78}\\
kLFDA \cite{Xiong_Gou_Camps:2014}&54.0  &\textbf{77.7}    &92.3\\
SCSP \cite{Chen_Yuan_Chen:2016}&57.3  &78.9   &91.5\\
\hline
Ours& \textbf{58.3} & 74.0 & 88.5  \\
\hline
\end{tabular}
\end{table}
\subsection{Analysis}
As observed in our experiments, hyper parameters $\mu$, $\gamma$ influence the performance of the proposed method. We give an empirical analysis in PRID2011 dataset in Section \ref{sec:paraan}. Note that the evaluation is conducted without temporal information. We also analyze the influence of different loss functions and the residual learning block, as well as the ranking results.

\subsubsection{Parameter Analysis}
\label{sec:paraan}

\begin{figure*}[htb]

\begin{minipage}[b]{.5\linewidth}
  \centering
  \centerline{\includegraphics[width=1\textwidth]{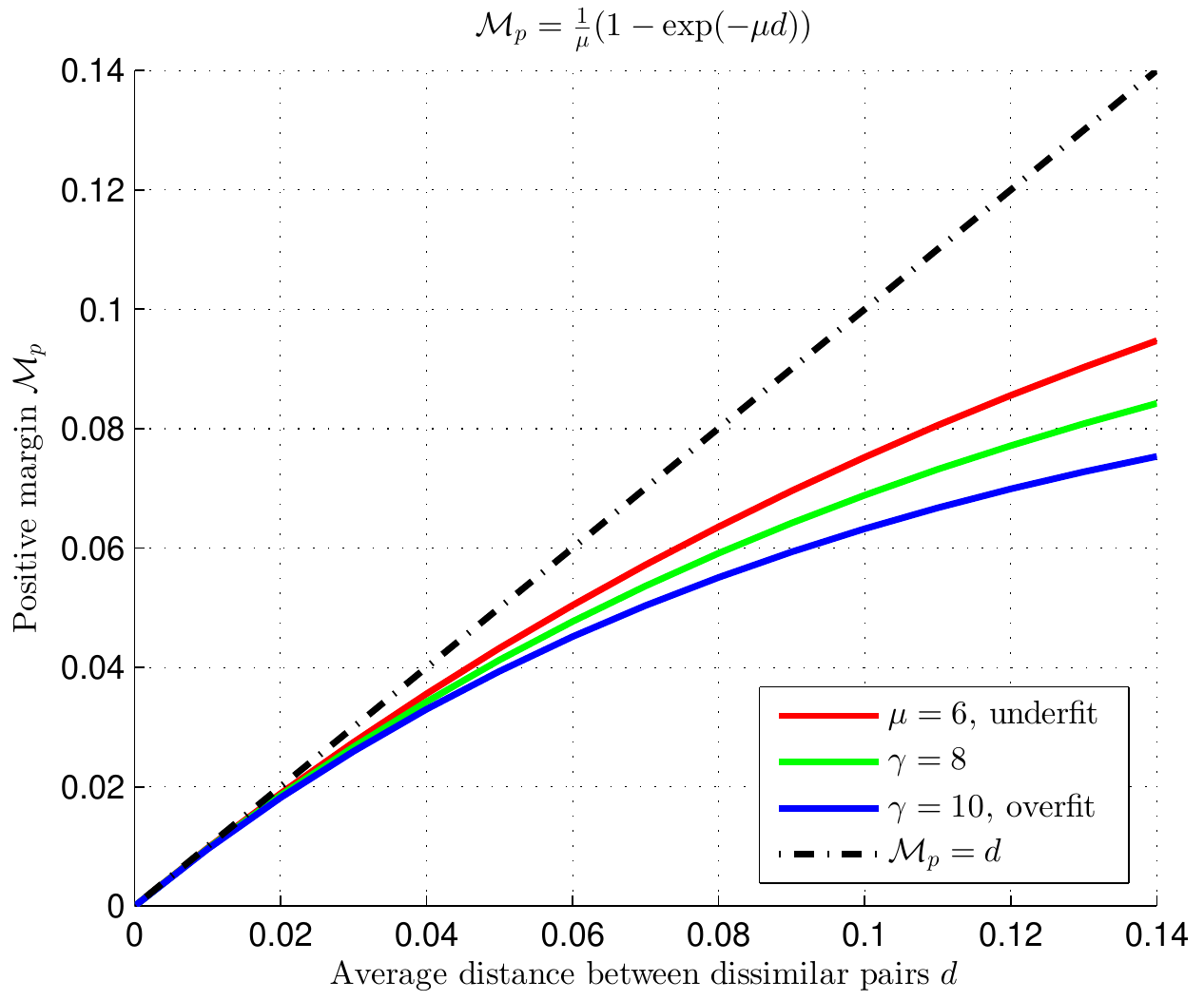}}
  \centerline{(a)}\medskip
\end{minipage}
\hfill
\begin{minipage}[b]{0.5\linewidth}
  \centering
  \centerline{\includegraphics[width=1\textwidth]{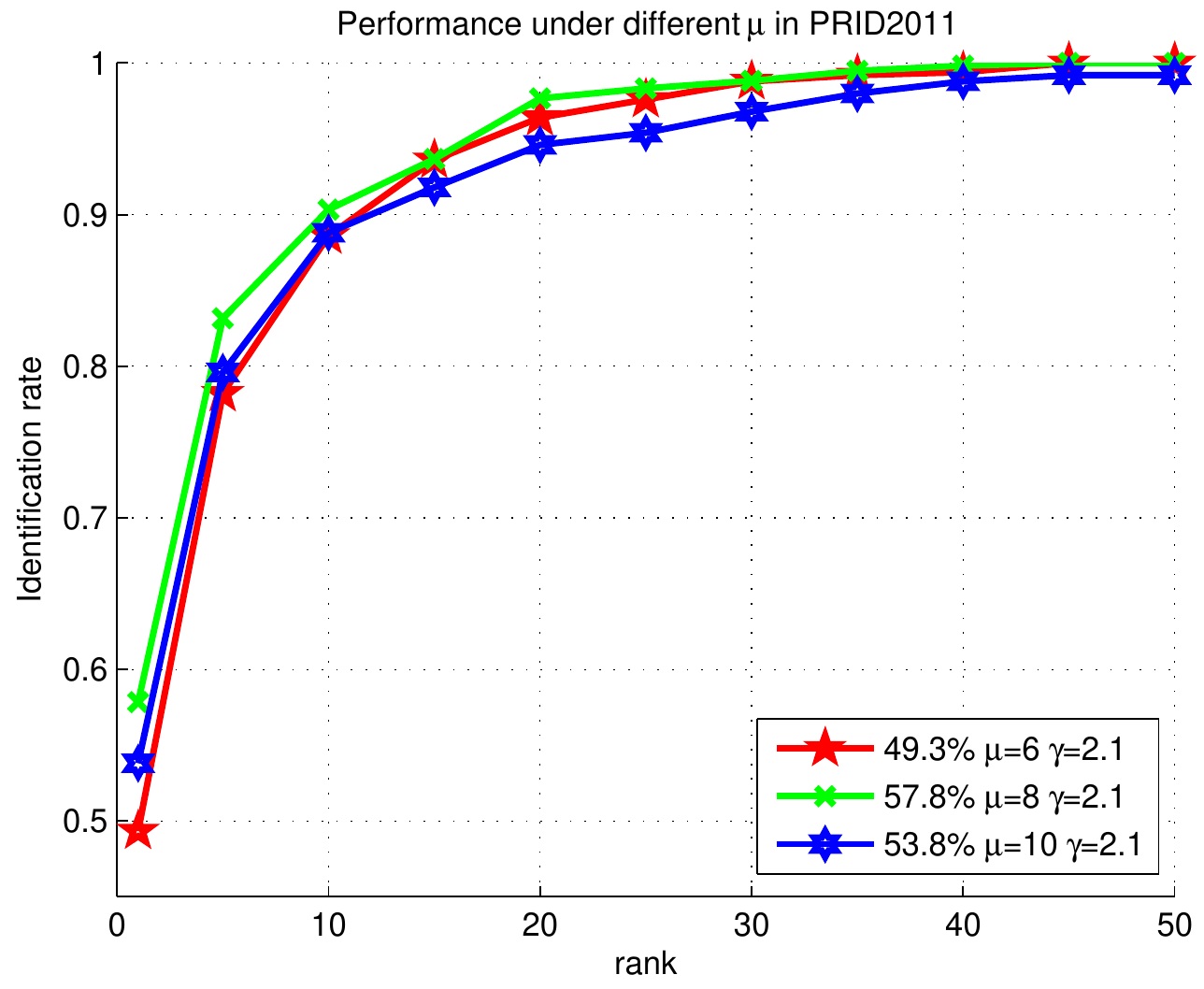}}
  \centerline{(b)}\medskip
\end{minipage}
\vfill
\begin{minipage}[b]{0.5\linewidth}
  \centering
  \centerline{\includegraphics[width=1\textwidth]{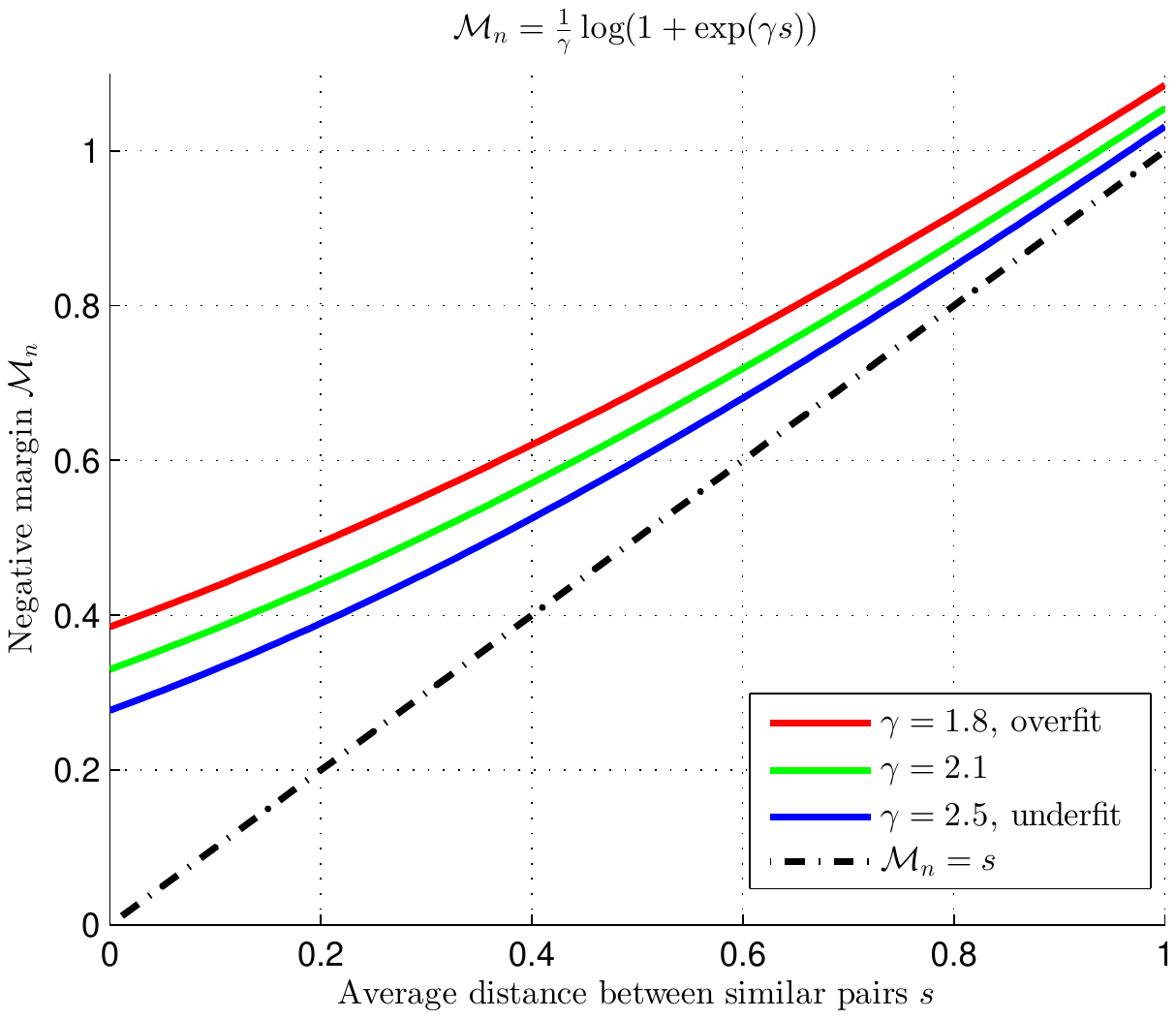}}
  \centerline{(c)}\medskip
\end{minipage}
\hfill
\begin{minipage}[b]{0.5\linewidth}
  \centering
  \centerline{\includegraphics[width=1\textwidth]{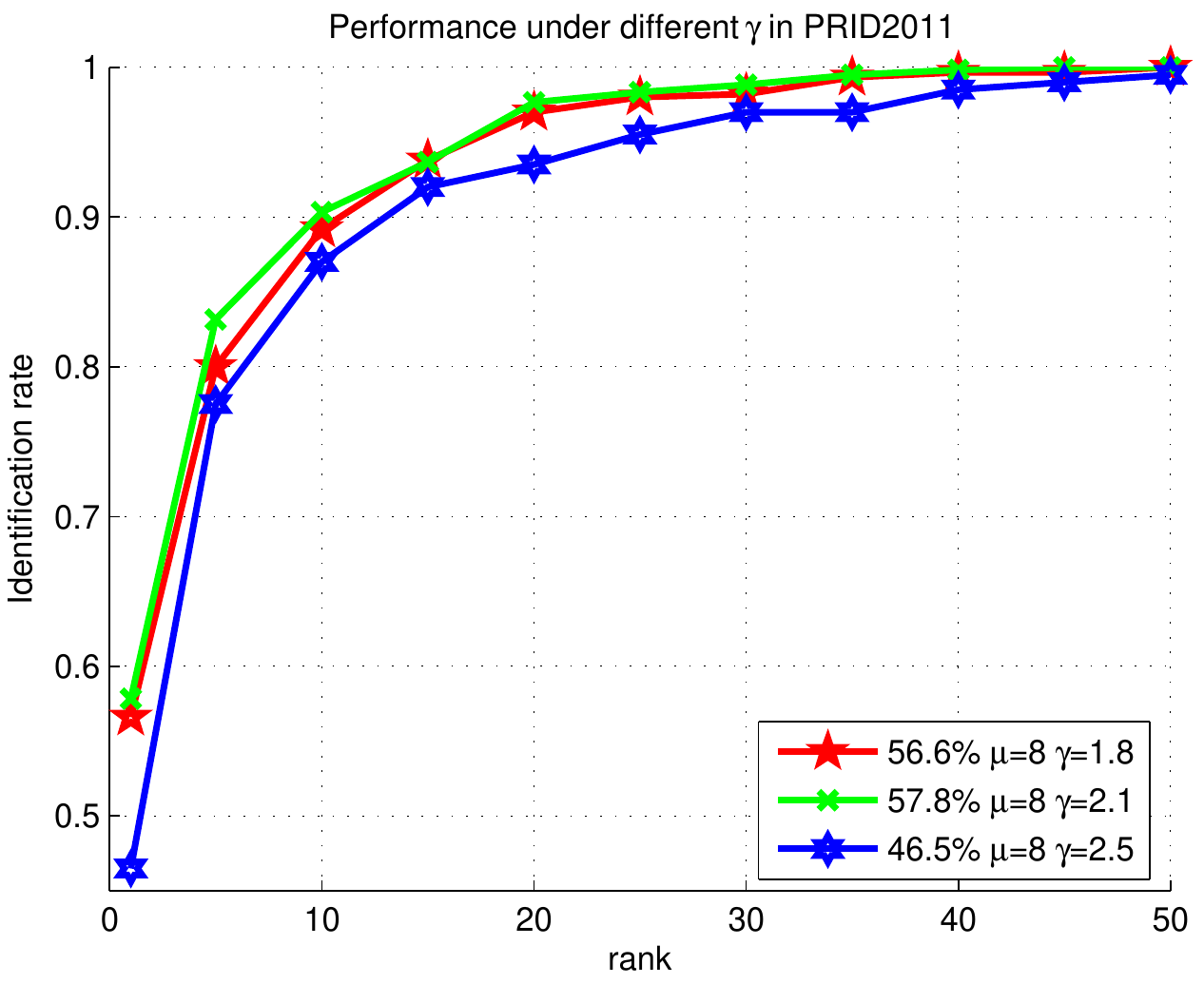}}
  \centerline{(d)}\medskip
\end{minipage}
\caption{The analysis of parameter on PRID2011 dataset. (a) illustration of the influence of parameter $\mu$: when $\mu=6, \mathcal{M}_p$ is rather substantial, the intra-class distance cannot be effectively narrowed, therefore leading to underfitting; when $\mu=10, \mathcal{M}_p$ is rather small, the intra-class distance is condensed too much, therefore leading to overfitting;  thus, choosing $\mu$ between 6 and 10 is ideal; (b) CMC curves with a fixed $\gamma$ and varying $\mu$, the deep architecture performs best when $\mu=8$; (c) illustration of the influence of parameter $\gamma$: when $\gamma=1.8, \mathcal{M}_n$ is rather big, the inter-class distance are driven to be too huge, therefore leading to overfitting; when $\gamma=2.5, \mathcal{M}_p$ is rather small, the inter-class distance cannot be effectively enlarged, therefore leading to underfitting; thus, choosing $\gamma$ between 1.8 and 2.5 is ideal; (d) CMC curves with a fixed $\mu$ and varying $\gamma$, the deep architecture performs best when $\gamma=2.1$.}
\label{fig:paracomp}
\end{figure*}

In Section \ref{sec:lam}, we introduce Eq. (\ref{eq_4}) as a nonlinear mapping method for calculating the average distance. 
As shown in Fig. \ref{fig_3}, hyper parameters $\mu, \gamma$ directly control the dynamic margin. We analyze the influence by changing one parameter while fixing the other one.

\textbf{The influence of $\mu$ on $\mathcal{M}_p$.} As depicted in Fig. \ref{fig_3},  $\mathcal{M}_p$ in Eq. (\ref{eq_4}) increases when $\mu$ decreases (from red, green to blue line). If $\mu$ is too huge, $\mathcal{M}_p$ can get extremely small, leading to overfitting. Similarly, too small $\mu$ gives huge down margin, making the constraint too weak and leads to underfitting.
The Fig. \ref{fig:paracomp}(b) plots the CMC curves with a fixed $\gamma$ and varying $\mu$ in PRID2011 dataset. The performance is optimal when $\mu=8$.

\textbf{The influence of $\gamma$ on $\mathcal{M}_n$.} When $\gamma$ increases (from blue, green to red line), $\mathcal{M}_n$ diminishes, as depicted in Fig. \ref{fig_3}. If $\gamma$ is too huge, underfitting might happen, due to small down-margin $\mathcal{M}_n$. Tiny $\gamma$ will lead to huge down-margin and overfitting. Results on PRID2011  with a fixed $\mu$ and varying $\gamma$ in Fig. \ref{fig:paracomp}(d) shows that, the performance reaches best when $\gamma=2.1$.

Although the identification rate will stay robust to a slight change in hyper parameters, tuning the 2 hyper parameters is necessary to avoid overfitting or underfitting, and  to get optimal perforamance.
,

\subsubsection{Analysis of loss functions and the residual block}


In this subsection, we evaluate the functions of two specific parts of the proposed method: i) the residual learning block in the proposed deep network; ii) the adaptive contrastive loss function. To review how each part contributes to the performance improvement, we implement five variants of the proposed person re-identification method, and compare their influence in the literature:

\textbf{Variant 1}: We replace the residual learning block with a convolution layer and use the conventional contrastive loss function to train the network. We refer this network as PCNN-C;

\textbf{Variant 2}: We replace the residual learning block with a convolution layer but retain the adaptive contrastive loss function. We refer this network as PCNN-AC;

\textbf{Variant 3}: We retain the residual learning block (as in Section \ref{sec:nn}, but only 1 residual block). However, conventional contrastive loss function is applied. We refer this network as RCNN-C;

\textbf{Variant 4}: We retain 1 residual learning block. However, conventional triplet loss function is applied. We refer this network as RCNN-T;

\textbf{Variant 5}: We retain 1 residual learning block as in {Section \ref{sec:nn} and adaptive contrastive loss function. We refer this network as RCNN-AC. This is our proposed integrative method.

The comparison results are shown in Table. \ref{table:moduleinf}. The following discussion mainly focus on the influence of two modules: adaptive contrastive loss function and residual learning block.

\begin{table}[!hbt]
\scriptsize
\centering
 \caption{ \label{table:moduleinf}
Comparison of module influence in terms of identification rates (\%) on different datasets}
\begin{tabular}{|c || c c||c || c c|}
\hline

Dataset & method &   $r=1$  & Dataset & method &   $r=1$  \\
\hline
\multirow{4}{*}{PRID2011}&PCNN-C  &48.5 &\multirow{4}{*}{Market1501}&PCNN-C &72.3   \\
 &PCNN-AC &52.4 & &PCNN-AC  &78.7   \\
 &RCNN-C  &53.8 & &RCNN-C &74.4 \\
  &RCNN-T &55.2 & &RCNN-T &76.6 \\
 &\textbf{RCNN-AC}  &\textbf{57.5} & &\textbf{RCNN-AC}  &\textbf{80.6}  \\
\hline
\hline
\multirow{4}{*}{CUHK01}&PCNN-C  &60.8 &\multirow{4}{*}{3DPeS}&PCNN-C  &41.7   \\
 &PCNN-AC &70.6 & &PCNN-AC  &52.8   \\
 &RCNN-C  &62.6 & &RCNN-C &46.7 \\
  &RCNN-T &65.0 & &RCNN-T &48.2 \\
 &\textbf{RCNN-AC}  &\textbf{71.9} & &\textbf{RCNN-AC}  &\textbf{58.3}  \\
\hline
\end{tabular}
\end{table}

\textbf{The influence of adaptive contrastive loss function}.
We compare the adaptive margin loss function with the conventional contrastive loss function and the triplet loss function, respectively. Compared with contrastive loss function, the top-1 identification rate of PRID2011, Market1501,CUHK01 and 3DPeS increases by 3.7\%, 6.2\%,  9.3\%, 11.6\% with residual model, and 3.9\%, 6.4\%,  9.8\%, 11.1\% without residual model, respectively. Compared with the triplet loss, the top-1 identification rate of PRID2011, Market1501,CUHK01 and 3DPeS increases by 2.3\%, 4.0\%,  6.9\%, 10.1\%, respectively. 


\textbf{The influence of residual learning block}.
On adopting residual learning block, the top-1 identification rate of PRID2011, Market1501,CUHK01 and 3DPeS increases by 5.2\%, 2.0\%,  1.6\%, 5.3\% on average, respectively.
Besides, the overall identification rate also improves on applying the residual learning block on the 4 datasets.

\textbf{The influence of combining 2 modules}.
On applying both the adaptive contrastive loss function and residual learning block, compared with conventional contrastive loss function, the top-1 identification rate of PRID2011, Market1501, CUHK01 and 3DPeS increases by 9.0\%, 8.3\%,  11.1\%, 16.6\% on average, respectively. 

We can reach the following 3 empirical conclusion from the experimental results: 1) Adaptive margin loss function performs better than conventional contrastive and triplet loss function; 2) Residual block contributes to the improvement of the method's performance; 3) Combining adaptive margin loss function and the proposed deep architecture achieves higher performance.

\subsubsection{Network Depth and transfer learning}
\label{sec:deeper}
In this subsection, we analyze the influence of the depth of the network and transfer learning, on the large-scale person re-identification dataset, Market 1501. Table \ref{table:deep} shows the experimental results, note that \textit{Res} stands for the residual learning block, and the proposed adaptive margin loss is used for all the experiments listed.

As shown in Fig. \ref{fig_4}, the number of the residual blocks can be changed. For small-sized and medium-sized dataset, namely PRID2011, CUHK01 and 3DPeS, we only use 1 residual block in the network. For large scale dataset Market1501, which contains 1501 identities, we evaluate the performance of our method with deeper network by adding residual blocks.

When the number of the residual learning block increases from 1 to 2, the top-1 accuracy improves 2.8 percent. Batch normalization \cite{ioffe2015batch} can accelerate deep network training by reducing internal covariate shift, as well as increasing the performance. When adding batch normalization to the residual learning block, the top-1 accuracy improves 1.5 percent for 2 residual blocks. We continue to make the network deeper, using more residual learning blocks with batch normalization, but the improvement becomes less significant  when it comes to 4 residual learning blocks. Overall, compared with only 1 residual learning block, we improve the accuracy by 6.4 percent,  by using 4 residual learning block with batch normalization. 

Geng et al. \cite{geng2016deep} points out that transfer learning will improve the performance for  person re-identification.  We also investigate the influence of transfer learning on the Market1501 dataset. Firstly, we train the deep network on another large scale person re-identification dataset, CUHK03 \cite{li2014deepreid} dataset. Next, we fine-tune it on our Market1501 dataset. The performance on 4 residual learning blocks with batch normalization improves 2.5 percent, and the final accuracy for Market1501 dataset comes to 89.5 percent.

\begin{table}[htb]
\centering
\caption{Network depth and fine-tune on Market 1501 (\textit{Res} stands for the residual learning block)}
\label{table:deep}
\scalebox{0.8}{\begin{tabular}{|c|c|c|c|c|c|c|}
\hline
               & 1 Res & 2 Res & 2 Res + BN & 3 Res + BN & 4 Res + BN & 4 Res + BN + fine-tune \\ \hline
Top-1 accuracy & 80.6  & 83.4  & 84.9       & 86.8       & 87.0       & 89.5                 \\ \hline
\end{tabular}}
\end{table}

\subsubsection{Ranking result analysis}
Some representative top-5 ranking results on the 4 datasets are shown in Fig. \ref{fig:ranking}. Most of the ranking results have high similarity, and some of them are even hard to discriminate by human eyes. According to \cite{wj31}, if we can emphasize more on narrowing the distance between the difficult negative references and its anchor, the accuracy of the ranking result may be enhanced further. In the future work, incorporating the mechanism of hard-negative strategy to our adaptive contrastive loss function may also be expected to improve the performance of the proposed method.
\begin{figure*}[htb]
\centering
\includegraphics[width=1.0\textwidth]{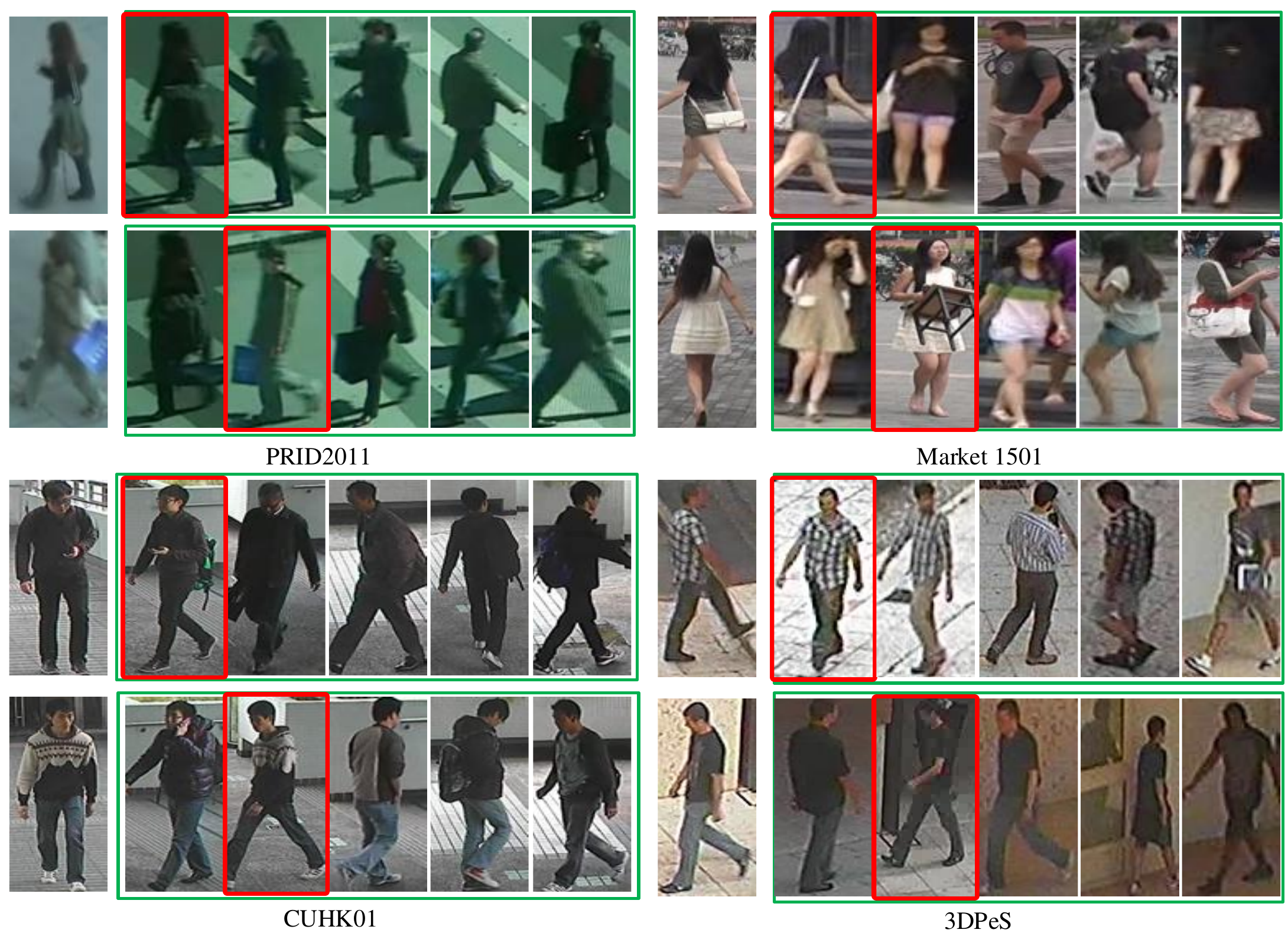}
\caption{Ranking result of PRID2011, Market1501, CUHK01 and 3dPeS. The left row is a query in test phase. The right block in green frame corresponds to the returned top5 results of the query. The one marked by red is the matched reference of the query.}
 \label{fig:ranking}
\end{figure*}

\section{Conclusion}
\label{sec_con}
In this paper, we present a novel method for person re-identification, which applies an adaptive margin strategy to learn a deep ranking model in the siamese framework. In order to learn discriminative and stable feature representations, we build a part-based deep neural network, in which different body parts are first discriminately learned in the global sub-network and local sub-network, and then fused in the fusion sub-network. The output features are further fed into the loss layer to penalize an adaptive margin between the positive pairs and negative pairs, so as to learn the deep model in an incremental manner. Experimental results on the PRID2011, Market1501, CUHK01 and 3DPeS datasets show that our method outperforms the state-of-the-art in person re-identification.

\section*{Acknowledgment}

This work is supported by the National Key Research and Development Program of China under Grant No. 2016YFB1001004, and the National Science Foundation of China under Grant No. 61473219.
\clearpage
\section*{References}
{\small
\bibliographystyle{elsarticle-num}
\bibliography{AM,refs}
}

\end{document}